\documentclass[preprint,12pt]{elsarticle}

\usepackage{amssymb}
\usepackage{bbm}
\usepackage{times}
\usepackage{epsfig}
\usepackage{graphicx}
\usepackage{amsmath}
\usepackage{amssymb}
\usepackage{multirow}
\usepackage{caption}
\usepackage{subcaption}
\usepackage[pagebackref=true,breaklinks=true,colorlinks,bookmarks=false]{hyperref}

\journal{Neurocomputing}

\begin{document}

\newcommand{\systemname}{Adv-4-Adv}
\newcommand{\tb}{\textbf}

\begin{frontmatter}

\title{Adv-4-Adv: Thwarting Changing Adversarial Perturbations via Adversarial Domain Adaptation}


\author[inst1]{Tianyue Zheng}

\affiliation[inst1]{organization={School of Computer Science and Engineering, Nanyang Technological University},
            addressline={50 Nanyang Ave}, 
            postcode={639798}, 
            country={Singapore}}

\author[inst1]{Zhe Chen}
\author[inst1]{Shuya Ding}
\author[inst1]{Chao Cai}
\author[inst1]{Jun Luo}


\begin{abstract}

Whereas \textbf{adversarial training} can be useful against specific adversarial perturbations, they have also been proven ineffective in generalizing towards attacks deviating from those used for training. However, we observe that this ineffectiveness is intrinsically connected to \textbf{domain adaptability}, another crucial issue in deep learning for which \textbf{adversarial domain adaptation} appears to be a promising solution.
Consequently, we propose \textbf{\systemname} as a novel adversarial training method that aims to retain robustness against unseen adversarial perturbations. 
Essentially, \systemname\ treats attacks incurring different perturbations as distinct domains, and by leveraging the power of adversarial domain adaptation, it aims to remove the domain/attack-specific features. This forces a trained model 
to learn a robust domain-invariant representation, which in turn enhances its generalizability. 
Extensive evaluations on Fashion-MNIST, SVHN, CIFAR-10, and CIFAR-100 demonstrate that a model trained by \systemname\ based on examples crafted by simple attacks (e.g., FGSM) can be generalized to more advanced attacks (e.g., PGD), and the performance exceeds state-of-the-art proposals on these datasets.

\end{abstract}




\end{frontmatter}

\section{Introduction}
As a key application of machine learning, image classification has witnessed a quantum leap in its performance since the boom of deep learning~\cite{krizhevsky2017imagenet,simonyan2014very, he2016deep, huang2017densely}. Nonetheless, researchers have recently discovered that this performance leap is built upon a shaky basis, as subtly but purposefully perturbed images (a.k.a.\ \textit{adversarial examples}) are shown to be able to cheat deep learning models so as to produce wrong class predictions~\cite{szegedy2013intriguing, goodfellow2014explaining, papernot2016limitations, moosavi2016deepfool}. 
What appears to be more disturbing is that these adversarial examples are inherently transferable across different models~\cite{tramer2017space, zhou2018transferable, naseer2019cross, papernot2016transferability}; this makes such attacks more effective while inevitably rendering the design of a universal defense mechanism extremely challenging.

Ever since the seminal discoveries on adversarial attacks~\cite{szegedy2013intriguing, goodfellow2014explaining, papernot2016limitations, moosavi2016deepfool}, various proposals have been made to thwart such attacks from different perspectives, including typically adversarial training (AT)~\cite{goodfellow2014explaining,kurakin2016adversarial,wong2018provable}, obfuscated gradient~\cite{buckman2018thermo,ma2018characterizing,dhillon2018stoc}, and various denoising techniques~\cite{meng2017magnet,song2018pixeldefend,xie2019denoise}.
Among these defense mechanisms, AT proposals~\cite{kurakin2016adversarial, madry2017towards} have proven very effective in thwarting specific attacks used to craft the training examples. Subsequent AT proposals further enhance the performance by encouraging the model to make low-confidence predictions on adversarial examples~\cite{stutz2020confidence}, utilizing the hypersphere embedding mechanism~\cite{pang2020boosting}, learning robust local features~\cite{song2019robust}, applying learned smoothing~\cite{chen2021robust}, injecting noisy labels during training~\cite{zhang2022noilin}, perturbing both image and label~\cite{wang2019bilateral}, and incorporating virtual adversarial examples~\cite{miyato2018virtual}. However, the ability of generalizing a given AT across different attacks can be rather questionable~\cite{kurakin2016adversarial,tramer2017ensemble}, thus substantially limiting the applicability of this method.

Different from conventional machine learning that demands handcrafted features~\cite{hastie09element}, a deep learning model extracts features from an input image automatically, which must be the key to adversarial vulnerability~\cite{ilyas2019adversarial}. As insightfully pointed out by Ilyas et al.~\cite{ilyas2019adversarial}, existing models largely rely on highly predictive (useful) but imperceptible (non-robust) features to achieve a high classification accuracy. As a result, purposefully perturbing these non-robust features can be visually imperceptible yet adversarially effective, and such adversarial perturbations are naturally transferable across models. In the meantime, proposals based on this observation have been made to remove non-robust features, by either regularizing the loss~\cite{zheng2016stability,kurakin2016adversarial,song2019improving} or resorting to robust optimization~\cite{wong2018provable,madry2017towards}. All these proposals claim a certain level of transferability across typical attacks, thanks to the common nature of such attacks.

However, as shown by both~\cite{tsipras2019robustness} and \cite{zhang2019theoretically}, a direct consequence of removing non-robust features is a degraded classification accuracy on clean images, simply due to the heavy reliance of a model on the useful but non-robust features~\cite{ilyas2019adversarial}. Whereas this inherent tradeoff appears to be inevitable, the ``brute-force" feature removing techniques (regularization~\cite{kurakin2016adversarial,song2019improving} or robust optimization~\cite{wong2018provable,madry2017towards}) may have exacerbated this situation, as the selection processed is defined by handcrafted functions quite against the spirit of deep learning. If one could find a more natural way of diminishing the impact of the non-robust features, it might improve the strength of defense while largely maintaining the classification accuracy on clean images.

\begin{figure}[t]
	\centering
	\includegraphics[width=.62\linewidth]{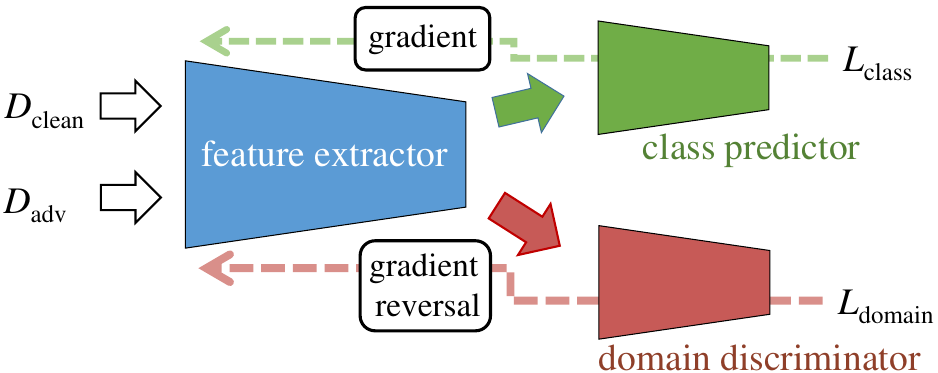}
	\caption{The basic idea of \systemname.}
	\label{fig:idea}
	\vspace{-2ex}
\end{figure}

To this end, we revisit the idea of \textit{domain adaptation} (DA)~\cite{shai2010domain} by considering the adversarial examples generated by a given attack as a special domain.
Instead of using the existing (loss-regularized) approach, we innovatively revise the \textit{adversarial domain adaptation} (ADA) framework~\cite{ganin2015unsupervised}, in order to suppress the impact of non-robust features in a ``soft'' manner. Essentially, we train a model that predicts (image) class labels using examples from two domains, namely clean (i.e., clean examples) and adversarial (i.e., adversarial examples), assisted by an additional discriminator that differentiates these two domains. While the predictor is trained to maintain the classification performance on both domains, our Adv-4-Adv method aims to ``cheat'' the discriminator by reversing the gradient during the backpropagation process, as illustrated in Figure~\ref{fig:idea}. This procedure suppresses only the non-robust features exploited (hence manipulated) by the adversarial domain, potentially striking a good balance between generalizable defense ability and clean image classification performance. 

We perform extensive evaluations of our Adv-4-Adv method on several standard benchmark datasets. We choose FGSM domain that contains adversarial examples crafted by~\cite{goodfellow2014explaining} for our training, and we examine the cross-domain performance of Adv-4-Adv on other adversarial domains induced by, for example, PGD~\cite{madry2017towards}. The experiment results have demonstrated perceivable improvements on both the generalization ability and the classification accuracy on clean data, under most circumstances. In the meantime, we also realize that fighting the ever-changing adversarial landscape is a very challenging objective, so our Adv-4-Adv is never meant to significantly outperform other, but rather to act as an alternative and offer a new perspective.

\section{Background and Related Work}
In this section, we discuss the basic notations and methodologies closely related to our design of Adv-4-Adv. In particular, we introduce typical adversarial attacks, their defense schemes, as well as a brief background on adversarial domain adaptation that has motivated our design.

\subsection{Adversarial Attack}
Arguably the first adversarial attack, Fast Gradient Sign Method (FGSM)~\cite{goodfellow2014explaining} is very simple to implement and hence still rather popular. FGSM generates adversarial examples by applying perturbations along an amplified (in $\ell_\infty$ sense) gradient direction as follows:
\begin{equation}
    \mathbf{x}^\mathrm{adv}=\mathbf{x}+\epsilon \cdot \operatorname{sign}\left(\nabla_\mathbf{x} \mathcal{L}\left(\mathbf{x}, \mathbf{y}^\mathrm{true}\right)\right), \label{eqn:fgsm}
\end{equation}
where $\mathbf{x}$ and $\mathbf{y}$ are the clean examples and their respective labels, $\epsilon$ is the strength of adversarial perturbations, $\operatorname{sign}$ is the signum function, $\nabla_\mathbf{x}$ is the gradient operator with respect $\mathbf{x}$, and $\mathcal{L}$ is the loss function for training. In order to improve the strength of the attack, Kurakin et al.~\cite{kurakin2016adversarial} later propose Projected Gradient Descent (PGD) as an enhanced version of FGSM with an iterative perturbation that is always clipped within $[-\epsilon, \epsilon]$ of the original pixels:
\begin{equation}
    \mathbf{x}^\mathrm{adv}_{n+1}=\operatorname{clip}\left(\mathbf{x}^\mathrm{adv}_n+ \alpha \cdot \operatorname{sign}\left(\nabla_\mathbf{x} \mathcal{L}\left(\mathbf{x}^\mathrm{adv}_n, \mathbf{y}^\mathrm{true}\right)\right)\right).
\end{equation}

Another strength enhancement to FGSM is R+FGSM~\cite{tramer2017ensemble}; it augments FGSM by adding a small randomized step as follows:
\begin{equation}
    \mathbf{x}^\mathrm{adv}=\mathbf{x}+\epsilon \cdot \operatorname{sign}\left(\nabla_\mathbf{x} \mathcal{L}\left(\mathbf{x}+\alpha \operatorname{sign}\left(\mathcal{N}(\mathbf{0},\mathbf{I})\right), \mathbf{y}^\mathrm{true}\right)\right),
    \label{eqn:fgsm}
\end{equation}
where $\alpha$ is a small random noise and $\mathcal{N}(\mathbf{0},\mathbf{I})$ represents a normal distribution with mean zero and identity covariance matrix. Given these basic attacks, several variants have been proposed later, including the Momentum Iterative FGSM (MIM for short)~\cite{dong2018boosting} that replaces the gradient with its accumulated version.



\subsection{Adversarial Defense}
Although adversarial training (AT)~\cite{goodfellow2014explaining,kurakin2016adversarial,wong2018provable} is not the only defense scheme, other schemes have been proven less effective. For example, obfuscated gradient~\cite{buckman2018thermo,ma2018characterizing,dhillon2018stoc} may give a false sense of security that can be readily circumvented~\cite{athalye2018false}, whereas denoising techniques~\cite{meng2017magnet,song2018pixeldefend,xie2019denoise} may not offer a stable performance across attacks. However, AT requires laboriously crafting a large number of adversarial examples, so we focus only on enhancing AT's generalization ability in low (adversarial) data regimes. In addition, as AT involves adversarial examples in training, it inevitably affects the classification accuracy on clean examples~\cite{tsipras2019robustness}. Whereas the tradeoff between (generalizable) robustness and (clean) accuracy can be clearly interpreted if robust optimization~\cite{wong2018provable,madry2017towards} is adopted as the defense scheme, such approaches may not guarantee the randomly sampled adversarial examples to be sufficiently representative~\cite{song2019improving}. Other training schemes either augment the datasets with cross-model examples~\cite{tramer2017ensemble} or regularize the loss~\cite{zheng2016stability,kurakin2016adversarial,song2019improving}. While involving cross-model examples in training may not offer a real generalization ability, regularizing loss to affect features (or model embeddings) can be too artificial and hence damage a desired tradeoff.

\subsection{From AT to Domain Adaptation}
Domain adaptation (DA)~\cite{shai2010domain,long2015dan,sun2016easy} is the ability to apply the classifier trained in a \textit{source domain} to different but related \textit{target domains}. Drawing inspiration from DA, one may clearly identify the analogy between DA and the generalization of AT: an \textit{adversarial domain} contains adversarial examples crafted by a certain attack. It might be 
feasible to adapt a model trained on a stronger domain (e.g., crafted by PGD) to a weaker domain (e.g., crafted by FGSM), but the key question here is: \textit{can we adapt a model trained on a weaker domain to stronger ones?} Although this can be achieved by regularization that enforces a domain-invariant embedding~\cite{song2019improving}, this method often relies on moments (e.g., mean or covariance) to measure the differences in (embedding) distributions, which can be insufficient given the diversity in distributions. On the contrary, the \textit{adversarial domain adaptation}\footnote{The term ``adversarial'' here stems from that of the generative adversarial network (GAN)~\cite{goodfellow2014gan}, whose meaning is quite different from that of AT.} (ADA)~\cite{ganin2015unsupervised,tzeng2017adversarial,shen2017wasserstein} offers a more natural way of enhancing the cross-domain generalization ability, because the process of assimilating the embedding distributions between two domains is implicitly driven by a trainable discriminator.

\section{\systemname: Adversarial Domain Adaptation to Counter Adversarial Perturbations} \label{sec:adv4adv}
We first explain the notations and define our problem. Then we present the basic version of \systemname, followed by the rationale behind our design, as well as its enhanced version for improved effectiveness. 

\subsection{Notations and Problem Definition}
We consider a typical network model $\mathbf{y} = g_{\boldsymbol{\theta}}(\mathbf{x}^\mathrm{clean})$ for image classification, with $\boldsymbol{\theta}$ denoting the model parameters, $\mathbf{x}^\mathrm{clean} = \{x^\mathrm{clean}_i\}$ as the clean examples sampled from a clean domain $\mathcal{D}^\mathrm{clean}$, and $\mathbf{y} = \{y_i\}$ as the corresponding true labels. To generate a certain adversarial domain $\mathcal{D}^\mathrm{adv}$, we craft its adversarial examples $\mathbf{x}^\mathrm{adv} = \{x^\mathrm{adv}_i\}$ via a known attack but we maintain the corresponding true labels. It is known that, although adversarial examples are visually indistinguishable from their corresponding clean examples, they may differ a lot in the internal representations of $g_{\boldsymbol{\theta}}$. Therefore, we consider two such representation spaces, namely i) \textit{feature} $\Phi$ expanded by the output of feature extractor (see Figure~\ref{fig:idea}) and ii) \textit{logit} $\Lambda$ appearing right before the final softmax function.

We intend to perform adversarial training in a DA manner, i.e., the training involves examples from both $\mathcal{D}^\mathrm{clean}$ and a particular $\mathcal{D}^\mathrm{adv}$ (preferably a weaker one generated by a simple attack such as FGSM), and the testing is done on any other domain generated by a different (stronger) attack. This DA is very different from the conventional ones that have plentiful labeled data in a source domain and aim to transfer the learned knowledge from the source domain to a target domain have none or few labeled data. In our case, we have two source domains, namely $\mathcal{D}^\mathrm{clean}$ and $\mathcal{D}^\mathrm{FGSM}$, both fully labeled, and the target domain, say $\mathcal{D}^\mathrm{PGD}$, used for testing is not only unlabeled but also offering no data to training. The reason is that the effort needs to be saved for AT is not labeling but example crafting; the labels are already offered by $\mathcal{D}^\mathrm{clean}$ for free. Formally, we require \systemname\ to train $g_{\boldsymbol{\theta}}$ on both $\mathcal{D}^\mathrm{clean}$ and $\mathcal{D}^\mathrm{FGSM}$, so as to achieve a generalizable performance on other unseen $\mathcal{D}^\mathrm{adv}$, while maintaining the performance on $\mathcal{D}^\mathrm{clean}$.

\subsection{Basic \systemname\ Architecture}
Given two datasets $\mathbf{x}^\mathrm{clean}$ and $\mathbf{x}^\mathrm{FGSM}$ sampled on $\mathcal{D}^\mathrm{clean}$ and $\mathcal{D}^\mathrm{FGSM}$ respectively\footnote{In the following, we fix the source adversarial domain as $\mathcal{D}^\mathrm{FGSM}$, because FGSM is the simplest known attack by far. However, our method should be applicable to other types of source domains.}, we adapt the ADA framework proposed by Ganin and Lempitskey~\cite{ganin2015unsupervised} to train the model $g_{\boldsymbol{\theta}}$. According to Figure~\ref{fig:idea}, $g_{\boldsymbol{\theta}}$ is separated into three components: feature extractor $g^\mathrm{f}$, class predictor $g^\mathrm{p}$, and domain discriminator $g^\mathrm{d}$, with their respective parameters $\boldsymbol{\theta}^\mathrm{f}$, $\boldsymbol{\theta}^\mathrm{p}$, and $\boldsymbol{\theta}^\mathrm{d}$. In training $g_{\boldsymbol{\theta}}$, we aim to minimize the following overall loss on $\mathbf{x} = \mathbf{x}^\mathrm{clean} \cup \mathbf{x}^\mathrm{FGSM}$:

\begin{eqnarray}
 \mathcal{L}(\boldsymbol{\theta}^\mathrm{f}, \boldsymbol{\theta}^\mathrm{P}, \boldsymbol{\theta}^\mathrm{d}) & = & \mathbb{E}_{\mathbf{x}^\mathrm{clean},\mathbf{y}}\left[ L^\mathrm{p}\left(g^\mathrm{p}(g^\mathrm{f}(\mathbf{x}^\mathrm{clean})), \mathbf{y} \right) \right] \nonumber \\
 & + & \mathbb{E}_{\mathbf{x}^\mathrm{FGSM},\mathbf{y}}\left[ L^\mathrm{p}\left(g^\mathrm{p}(g^\mathrm{f}(\mathbf{x}^\mathrm{FGSM})), \mathbf{y} \right) \right] \nonumber \\
 & - & \beta \mathbb{E}_{\mathbf{x},\mathbf{d}}\left[ L^\mathrm{d}\left(g^\mathrm{d}(g^\mathrm{f}(\mathbf{x})), d \right)\right], \label{eqn:loss}
\end{eqnarray}

where $\mathbf{d} = \{d_i\}$ contains the domain labels telling whether $x_i$ belongs to $\mathbf{x}^\mathrm{clean}$ or $\mathbf{x}^\mathrm{FGSM}$, $\beta$ is a regularization parameter for striking a good balance between the predictor loss $L^\mathrm{p}$ and the domain loss $L^\mathrm{d}$, and the minus sign is obtained by reversing the gradient during the backpropagation. The training process aims to reach a \textit{saddle point} of $\mathcal{L}(\boldsymbol{\theta}^\mathrm{f}, \boldsymbol{\theta}^\mathrm{P}, \boldsymbol{\theta}^\mathrm{d})$ by properly tuning the parameters:
\begin{equation}
    \begin{split}
     (\hat{\boldsymbol{\theta}}^\mathrm{f}, \hat{\boldsymbol{\theta}}^\mathrm{p}) & =  \arg\min_{\boldsymbol{\theta}^\mathrm{f}, \boldsymbol{\theta}^\mathrm{p}} \mathcal{L}(\boldsymbol{\theta}^\mathrm{f}, \boldsymbol{\theta}^\mathrm{P}, \boldsymbol{\theta}^\mathrm{d}) \\
     \hat{\boldsymbol{\theta}}^\mathrm{d} & =  \arg\max_{\boldsymbol{\theta}^\mathrm{d}} \mathcal{L}(\boldsymbol{\theta}^\mathrm{f}, \boldsymbol{\theta}^\mathrm{P}, \boldsymbol{\theta}^\mathrm{d})
    \end{split}
\end{equation}
While the functions of $\hat{\boldsymbol{\theta}}^\mathrm{p}$ and $\hat{\boldsymbol{\theta}}^\mathrm{d}$ are pretty clear: they minimize their respective classification losses, that of $\hat{\boldsymbol{\theta}}^\mathrm{f}$ is rather tricky. On one hand, $\hat{\boldsymbol{\theta}}^\mathrm{f}$ aims to (naturally) minimize the class prediction loss for both domains. On the other hand, it maximizes the domain discrimination loss, which can be interpreted with the \textit{non-robust feature} concept introduced by Ilyas et al.~\cite{ilyas2019adversarial}. According to the analysis in~\cite{ilyas2019adversarial}, the reason for $\mathbf{x}^\mathrm{FGSM}$ to become adversarial examples is that FGSM subtly tunes the non-robust but predictive features so as to cheat a normal predictor, so it is these features that discriminate the two domains $\mathcal{D}^\mathrm{clean}$ and $\mathcal{D}^\mathrm{FGSM}$. Consequently, the chance that $\hat{\boldsymbol{\theta}}^\mathrm{f}$ can maximize the domain discrimination loss is that such non-robust features are largely suppressed in the output of $g^\mathrm{f}$, resulting in  domain-invariant and hence robust features generalizable to other domains.

\begin{figure*}[t]
	\centering
	\includegraphics[width=.9\linewidth]{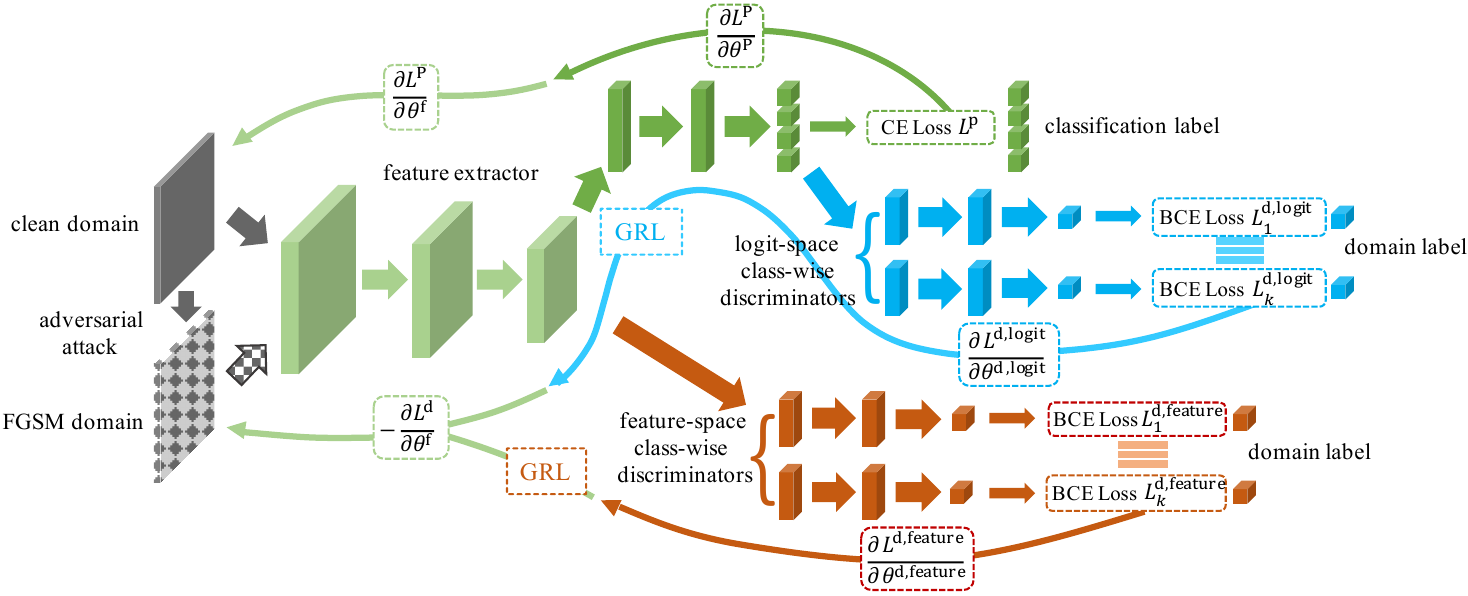}
	\caption{The architecture of \systemname, where CE and BCE represent cross-entropy and binary cross-entropy, respectively.}
	\label{fig:sysdiag}
\end{figure*}

\subsection{Multi-Representation \& -Class Enhancement}\label{ssec:representation_class}
As the output of $g^\mathrm{f}$ may fall into a very high-dimensional space, tuning it with only one loss function can lead to a very slow convergence. Therefore, \systemname\ requires two major enhancements to improve its effectiveness.

\paragraph{Multi-Representation Assisted DA} 
Although the feature space $\Phi$ can have a very high dimension, the logit space $\Lambda$ can be of a much lower dimension, while still being closely correlated with $\Phi$. Therefore, it makes sense to add another domain discriminator to the logits layer of the predictor, creating another ``backpropagtion force'' to enforce the robust and domain-invariant features. Essentially, we upgrade the basic loss in Eqn.~\eqref{eqn:loss} by adding the fourth term as follows:
\begin{eqnarray}\label{eq:represetation}
\mathcal{L}(\boldsymbol{\theta}^\mathrm{f}, \boldsymbol{\theta}^\mathrm{P}, \boldsymbol{\theta}^\mathrm{d}) & = & \mathbb{E}_{\mathbf{x}^\mathrm{clean},\mathbf{y}}\left[ L^\mathrm{p}\left(g^\mathrm{p}(g^\mathrm{f}(x^\mathrm{clean})), y \right) \right] \nonumber \\
 & + & \mathbb{E}_{\mathbf{x}^\mathrm{FGSM},\mathbf{y}}\left[ L^\mathrm{p}\left(g^\mathrm{p}(g^\mathrm{f}(x^\mathrm{FGSM})), y \right) \right] \nonumber \\
 & - & \beta \mathbb{E}_{\mathbf{x},\mathbf{d}}\left[ L^\mathrm{d}\left(g^\mathrm{d}(g^\mathrm{f}(x)), d \right)\right] \nonumber \\
 & - & \gamma \mathbb{E}_{\mathbf{x},\mathbf{d}}\left[ L^\mathrm{d}\left(\bar{g}^\mathrm{d}(g^\mathrm{p}(g^\mathrm{f}(x))), d \right)\right],
\end{eqnarray}
where $\bar{g}^\mathrm{d}$ is another domain discriminator that takes the logits of the class predictor $g^\mathrm{p}$ as its input, while $\beta$ and $\gamma$ are used to control the weights of these discriminators.

\paragraph{Multi-Class Enhanced DA} 
One major issue with directly applying a discriminator to either feature or logit space is that any differences between the two distributions can be taken as hint for discrimination, possibly leading to a false alignment between them. In fact, the discriminative structures related to individual classes should be preserved for the sake of accurate classification, but simply ``cheating'' (with reversed gradients) a discriminator to assimilate the two domains can potentially mix up the respective discriminative structures, damaging the classification accuracy. Therefore, we adopt the \textit{multi-domain adaptation} (MDA) approach~\cite{pei2018multi} to further enhance the training structure. Essentially, we associate each class $k$ with a feature discriminator $g^\mathrm{d}_k$ and a logit discriminator $\bar{g}^\mathrm{d}_k$, so the final loss appears as follows:
\begin{align} \label{mr_mc}
 \mathcal{L}(\boldsymbol{\theta}^\mathrm{f}, \boldsymbol{\theta}^\mathrm{P}, \boldsymbol{\theta}^\mathrm{d})&=  \mathbb{E}_{\mathbf{x}^\mathrm{clean},\mathbf{y}}\left[ L^\mathrm{p}\left(g^\mathrm{p}(g^\mathrm{f}(x^\mathrm{clean})), y \right) \right] \nonumber \\
 \!\!\!& +  \mathbb{E}_{\mathbf{x}^\mathrm{FGSM},\mathbf{y}}\left[ L^\mathrm{p}\left(g^\mathrm{p}(g^\mathrm{f}(x^\mathrm{FGSM})), y \right) \right] \nonumber \\
 \!\!\!& -  \beta \sum_{k=1}^K \mathbbm{1}_{y=k} \mathbb{E}_{\mathbf{x},\mathbf{d}}\left[ L^\mathrm{d}\left(g^\mathrm{d}_k (g^\mathrm{f}(x)), y, d \right)\right] \nonumber \\
 \!\!\!& -  \gamma \sum_{k=1}^K \mathbbm{1}_{y=k} \mathbb{E}_{\mathbf{x},\mathbf{d}}\left[ L^\mathrm{d}\left(\bar{g}^\mathrm{d}_k (g^\mathrm{p}(g^\mathrm{f}(x))), y, d \right)\right]\!\!,
\end{align}

\paragraph{\systemname\ DA} To summarize, we use Figure~\ref{fig:sysdiag} to illustrate the final architecture of our proposed \systemname\ training method. Given clean and adversarial examples $\mathbf{x}^\mathrm{clean}$ and $\mathbf{x}^\mathrm{adv}$ sampled from two domains as inputs,
a convolutional neural network $g^\mathrm{f}$ is employed to extract features from the images. After that, the upper branch consists of a class predictor $g^\mathrm{p}$ that transforms the embeddings from feature space to logit space, and predicts class labels. The middle branch is a set of binary class-wise discriminators $\bar{g}^\mathrm{d}$, leveraging the output of $g^\mathrm{p}$ to differentiate the two domains. The lower branch is another set of binary class-wise discriminators $g^\mathrm{d}$, applying to the feature space so as to reinforce the domain differentiation. Both sets of discriminators are prepended with Gradient Reversal Layers (GRL) to reverse gradient during backpropagation. By minimizing the sum of the cross-entropy loss of class predictors and the binary cross-entropy loss of domain discriminators (negated by GRL) as in Eqn.~\eqref{mr_mc}, \systemname\ adversarially trains a robust classifier whose defense capabilities can be generalized to 
unseen attacks.

\section{Experiments}
In this section, we evaluate \systemname\ in both white-box and black-box settings. By comparing \systemname\ with state-of-the-art adversarial training methods, we demonstrate its capability in defending against unseen attacks, while keeping a satisfactory accuracy for clean examples. We also visualize and quantify how the representation changes when we apply stronger attacks to the trained classifiers, confirming \systemname's ability in learning a robust domain-invariant representation and hence explaining its enhanced generalization ability. We then conduct an abalation study to justify our design rationale. Finally, we show \systemname\ can also be flexibly trained by other attacks.




\subsection{Datasets, Baselines and Implementation Details}

Four popular datasets are chosen to evaluate the performance of \systemname: Fashion-MNIST~\cite{xiao2017fashion}, SVHN~\cite{netzer2011reading}, CIFAR-10, and CIFAR-100~\cite{krizhevsky2009learning}. 
The model is trained on 60K Fashion-MNIST, 73K SVHN, 50K CIFAR-10, or 50K CIFAR-100 images. For convenience of comparison, the range of the pixel values of all datasets are scaled to $[0, 1]$.
We compare \systemname\ with the following baselines:

%
\begin{itemize}
  \item Normal Training (NT): the classifier is trained by minimizing the cross-entropy loss over clean images. 
  \vspace{-1ex}
  \item Standard Adversarial Training (SAT): the classifier is trained by minimizing the cross-entropy loss over clean images and adversarial images crafted for the model by the FGSM attack. 
  \vspace{-1ex}
  \item Ensemble Adversarial Training (EAT)~\cite{tramer2017ensemble}: the classifier is trained by minimizing the cross-entropy loss over clean images, and three sets of adversarial images crafted for the current model, and two other pre-trained models also by the FGSM attack.
  \vspace{-1ex}
  \item Adversarial Training with Domain Adaptation (ATDA)~\cite{song2019improving}: the classifier is trained by minimizing the sum of several loss functions over clean and adversarial images crafted for the model by FGSM, aiming to enforce a domain-invariant representation.
\end{itemize}
The magnitude of perturbations for FGSM attacks used for training the baseline methods are 0.1, 0.02, 4/255, and 4/255 in the $l_\infty$ norm for Fashion-MNIST, SVHN, CIFAR-10, and CIFAR-100, respectively.



We choose the base classifiers $g_{\boldsymbol{\theta}}$ to be same as that in~\cite{song2019improving}, so as to facilitate a direct comparison with this most related proposal. For white-box attacks, 
the classifier consists of a normalization layer for preprocessing, a series of convolutional and activation layers for feature extraction, and linear layer 
as the last layer for classification. For black-box attacks, we adopt a surrogate model~\cite{song2019improving} held out during the training process.

After initializing the classifiers by pretraining on clean images, we further adapt them to the adversarial domain by training with FGSM-crafted images, as introduced in Section~\ref{sec:adv4adv}. In the adaptation step, we have two variants, \systemname\ and \systemname* with different loss functions. In \systemname, the cross-entropy losses of both the clean and adversarial images, and the two sets of domain losses are all taken into account. Since the classifier is already pretrained with the clean images, in \systemname*, we only minimize the cross-entropy loss of adversarial images and the domain losses, while the cross-entropy loss of clean images is not considered. To prevent discriminators from removing extra robust features, we employ early stopping to stop training at the point when there is no performance improvement on the adversarial domain. 
%
We train the model for 400 epochs, use a batch size of 256, and use an Adam optimizer with a learning rate of 0.01 and betas of (0.9, 0.999). All experiments are implemented on Tesla V100 GPU.

\subsection{Performance Comparison}
We evaluate the performance of \systemname\ on the aforementioned datasets and compare 
with the baselines in both white-box and black-box settings. 
%
\begin{table}[h!]
    \scriptsize
    \setlength{\tabcolsep}{2pt}
    \caption{The accuracy of six defense methods on the testing datasets and the adversarial examples generated by various adversaries, where we abbreviate \systemname\ to A-4-A for brevity's sake.}
    
    \begin{subtable}[h]{\textwidth}
    \centering
    \caption{On Fashion-MNIST. The magnitude of perturbations is 0.1 in $\ell_{\infty}$ norm.}
    \label{stab:fmnist}
    \begin{tabular}{@{\extracolsep{4pt}}lcrrrrrrrrr@{}}
    \hline
     \multirow{2}{*}{Defense}& Clean  & \multicolumn{4}{c}{White-Box Attack (\%)} & \multicolumn{4}{c}{Black-Box Attack (\%)} \\
     \cline{3-6} \cline{7-10}
      & (\%) & FGSM & PGD-20 / 40 & R+FGSM & MIM & FGSM & PGD-20 / 40 & R+FGSM & MIM \\
     \hline
     NT & \tb{89.85} & 6.51 & 0.03 / 0.00 & 8.08 & 0.25 & 46.76 & 52.87 / 53.48 & 50.39 & 49.78 \\
     SAT & 83.78 & 60.02 & 13.27 / 10.72 & 58.59 & 26.82 & 68.25 & 69.67 / 70.00 & 69.72 & 69.69 \\
     EAT & 75.42 & 59.74 & 23.80 / 22.41 & 57.08 & 32.64 & 67.68 & 67.81 / 68.38 & 68.03 & 68.62 \\
     ATDA  & 75.95 & 56.91 & 16.47 / 14.19 & 55.79 & 27.76 & 69.06 & 69.45 / 69.53 & 68.80 & 68.37 \\ 
     A-4-A & 78.05 & 68.00 & 56.21 / 54.42 & 68.33 & \tb{59.76} & \tb{73.01} & \tb{74.74} / \tb{75.20} & \tb{74.00} & \tb{74.53}\\
     A-4-A* & 74.15 & \tb{68.83} & \tb{57.56 / 56.40} & \tb{69.73} & 58.18 & 72.40 & 73.10 / 72.98 & 72.54 & 72.65\\
     \hline
    \end{tabular}
	\end{subtable}

    \vspace*{0.2 cm}
    
    \begin{subtable}[h]{\textwidth}
    \centering
    \caption{On SVHN. The magnitude of perturbations is 0.02 in $\ell_{\infty}$ norm.}
    \label{stab:svhn}
    \begin{tabular}{@{\extracolsep{4pt}}lcrrrrrrrrr@{}}
    \hline
     \multirow{2}{*}{Defense}& Clean  & \multicolumn{4}{c}{White-Box Attack (\%)} & \multicolumn{4}{c}{Black-Box Attack (\%)} \\
     \cline{3-6} \cline{7-10}
      & (\%) & FGSM & PGD-20 / 40 & R+FGSM & MIM & FGSM & PGD-20 / 40 & R+FGSM & MIM \\
     \hline
     NT & \tb{84.71} & 24.76 & 8.74 / 6.96 & 27.66 & 14.37 & 68.66 & 74.88 / 75.08 & 70.53 & 69.61 \\
     SAT & 84.67 & 55.03 & 41.69 / 39.68 & 57.80 & 49.20 & 76.94 & 79.39 / 79.51 & 77.84 & \tb{77.53} \\
     EAT & 81.01 & 55.92 & 48.02 / 46.79 & 58.18 & 52.57 & 75.71 & 77.57 / 77.51 & 76.69 & 76.00 \\
     ATDA  & 81.63 & 54.60 & 47.00 / 45.29 & 57.40 & 51.43 & 74.44 & 76.53 / 76.61 & 75.33 & 74.39 \\ 
     A-4-A & 82.68 & \tb{57.11} & 48.36 / 47.03 & \tb{59.42} & 53.49 & \tb{78.06} & \tb{79.81 / 80.31} & \tb{81.87} & 77.48\\
     A-4-A* & 78.74 & 57.08 & \tb{51.04 / 49.76} & 59.39 & \tb{54.82} & 76.13 & 78.04 / 78.46 & 80.62 & 76.52  \\ 
     \hline
    \end{tabular}
	\end{subtable}

    \vspace*{0.2 cm}
    
	\begin{subtable}[h]{\textwidth}
    \centering
    \caption{On CIFAR-10. The magnitude of perturbations is 4/255 in $\ell_{\infty}$ norm.}
    \label{stab:cifar10}
    \begin{tabular}{@{\extracolsep{4pt}}lcrrrrrrrrr@{}}
    \hline
     \multirow{2}{*}{Defense}& Clean  & \multicolumn{4}{c}{White-Box Attack (\%)} & \multicolumn{4}{c}{Black-Box Attack (\%)} \\
     \cline{3-6} \cline{7-10}
      & (\%) & FGSM & PGD-20 / 40 & R+FGSM & MIM & FGSM & PGD-20 / 40 & R+FGSM & MIM \\
     \hline
     NT & \tb{85.74} & 11.47 & 0.42 / 0.22 & 13.13 & 1.51 & 59.77 & 52.80 / 50.63 & 62.74 & 54.84 \\
     SAT & 82.49 & 53.31 & 51.68 / 51.01 & 56.11 & 52.12 & 79.16 & 79.31 / 79.15 & 79.31 & 79.15 \\
     EAT & 78.77 & 56.00 & 54.71 / 54.15 & 58.15 & 55.25 & 76.69 & 77.46 / 76.91 & 77.26 & 77.16 \\
     ATDA  & 82.78 & 54.24 & 52.23 / 51.66 & 57.62 & 53.17 & 79.07 & 79.13 / 79.25 & 79.22 & 79.21 \\ 
     A-4-A & 81.44 & 57.73 & 56.35 / 56.52 & 59.63 & 57.37 & \tb{80.07} & \tb{80.27 / 80.70} & \tb{80.60} & \tb{80.77} \\
     A-4-A* & 79.88 & \tb{59.98} & \tb{59.05 / 58.81} & \tb{61.68} & \tb{59.56} & 78.65 & 78.93 / 79.28 & 79.19 & 78.86 \\
     \hline
    \end{tabular}
	\end{subtable}

    \vspace*{0.2 cm}
    
	\begin{subtable}[h]{\textwidth}
    \centering
    \caption{On CIFAR-100. The magnitude of perturbations is 4/255 in $\ell_{\infty}$ norm.}
    \label{stab:cifar100}
    \begin{tabular}{@{\extracolsep{4pt}}lcrrrrrrrrr@{}}
    \hline
     \multirow{2}{*}{Defense}& Clean  & \multicolumn{4}{c}{White-Box Attack (\%)} & \multicolumn{4}{c}{Black-Box Attack (\%)} \\
     \cline{3-6} \cline{7-10}
      & (\%) & FGSM & PGD-20 / 40 & R+FGSM & MIM & FGSM & PGD-20 / 40 & R+FGSM & MIM \\
     \hline
     NT & \tb{58.24} & 2.25 & 0.30 / 0.22 & 2.79 & 0.48 & 41.75 & 40.40 / 39.86 & 43.31 & 40.56 \\
     SAT & 55.66 & 26.69 & 24.57 / 24.42 & 28.93 & 25.84 & \tb{51.85} & \tb{51.92} / 51.65 & \tb{52.06} & 51.81 \\
     EAT & 52.07 & 28.99 & 28.03 / 27.86 & 31.02 & 28.14 & 49.90 & 50.26 / 50.21 & 50.04 & 49.69 \\
     ATDA  & 55.26 & 26.35 & 24.29 / 23.98 & 28.14 & 25.40 & 51.84 & 51.21 / 51.88 & 51.81 & 51.37 \\ 
     A-4-A & 52.41 & 29.44 & 28.72 / 29.15 & 31.33 & 29.54 & 51.30 & 51.79 / \tb{51.97} & 51.59 & \tb{51.83} \\
     A-4-A*  & 51.84 & \tb{31.31} & \tb{31.67 / 31.09} & \tb{33.48} & \tb{31.35} & 50.92 & 50.95 / 50.82 & 51.48 & 40.74   \\
     \hline
    \end{tabular}
	\end{subtable}
	\label{tab:accuracy}
	\vspace{-2ex}
\end{table}

The evaluation results are shown in Table~\ref{tab:accuracy}. The goal here is not to achieve state-of-the-art classification performance under specific attacks, but to showcase the generalization ability of our approach in terms of robustness across unseen attacks. As expected, NT achieves the highest accuracy on clean data, but fails all white-box adversarial attacks on all datasets, and its accuracy drops severely under black-box attacks. SAT has a decent performance under white-box FGSM attacks, but it does not generalize well to other white-box attacks. EAT, as the enhanced version of SAT, generalizes better to PGD, R+FGSM, and MIM attacks, at the cost of a reduced clean image classification accuracy. Although ATDA aims to improve the performance of EAT via DA, our results do not seem to confirm that.\footnote{The disagreement between our ATDA results and those reported in~\cite{song2019improving} may be explained by different learning libraries (Pytorch rather than TensorFlow), hyper-parameters, and data augmentation schemes.} 

It is our proposal, \systemname, that actually strikes a good balance between robustness and accuracy: it performs the best in all black-box defense tasks, and its variant, \systemname*, generalizes well and excels at white-box defense tasks, at a minor cost of accuracy on clean data. The difference between \systemname\ and \systemname* may be attributed to the fact that the cross-entropy loss of sole adversarial examples employed by \systemname* only provide defense against attacks for a specific model, while the addition of cross-entropy loss of clean examples add generalizability to black-box attacks whose model is unknown during training. We believe that the domain-invariant features induced by domain discriminators have largely contributed to the good performance of \systemname.

\subsection{Further Analysis of \systemname}

To further demonstrate the cross-attack defense capabilities of \systemname, we first visualize and compare the logit distributions among all comparison parties, then we prove the superiority of \systemname\ over other methods by analyzing the logit-space embeddings quantitatively.  

\begin{figure}[t]
	\centering
	\includegraphics[width=0.8\linewidth]{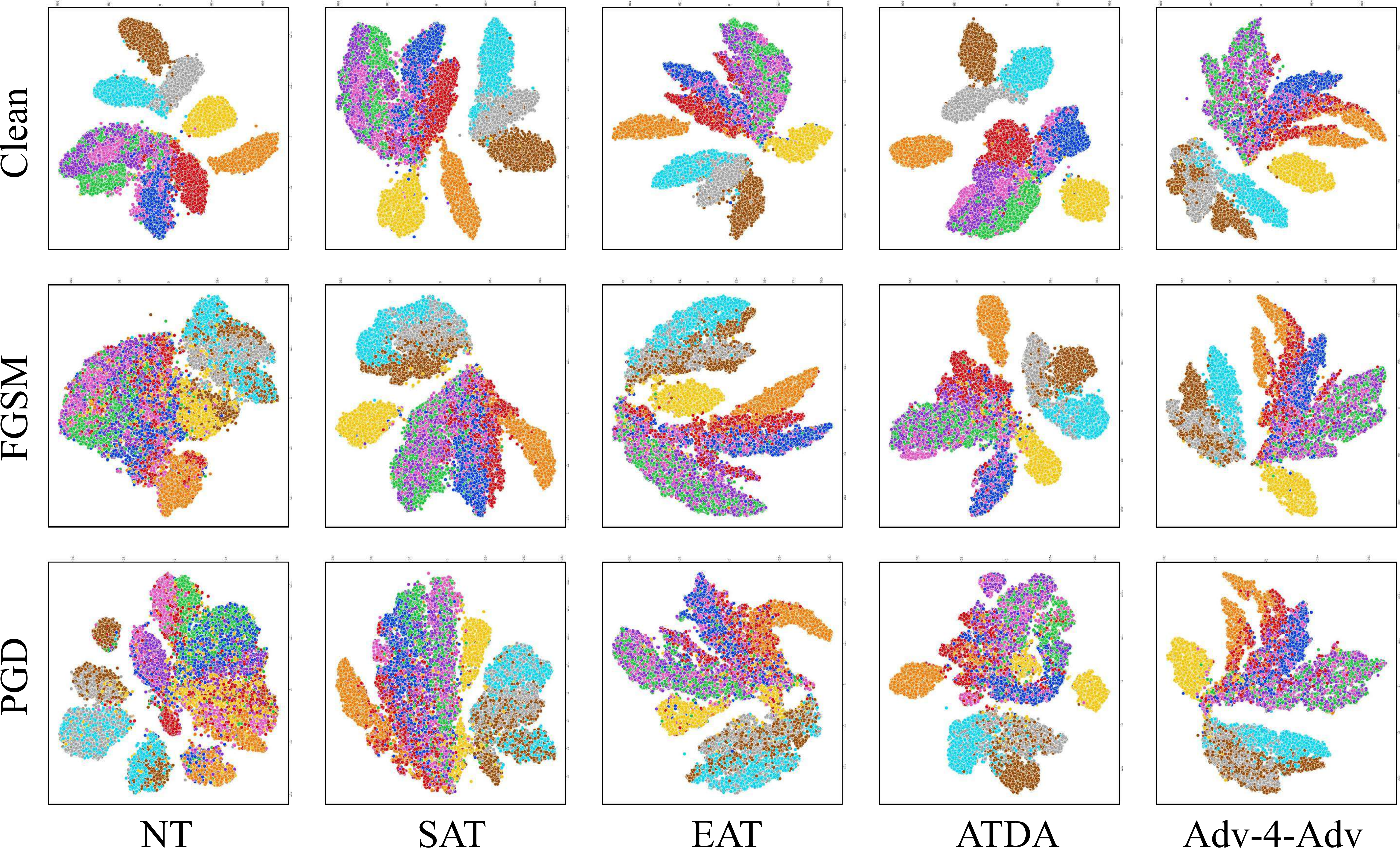}
	\caption{t-SNE visualizations of logit embeddings; they confirm that \systemname\ indeed generates domain-invariant representations.}
	\label{fig:tsne}
	\vspace{-2ex}
\end{figure}

\paragraph{Distribution Discrepancy}
We visualize the logit embeddings of the images using t-SNE~\cite{maaten2008visualizing}. The logit space representations of the images are projected onto a 2-D plane, and individual classes are colored differently, as shown in Figure~\ref{fig:tsne}. 
%
%

We observe a strong correspondence between the success 
in generalizing the classification accuracy of the defense methods (shown in Table~\ref{tab:accuracy}) and the consistency of the shapes of t-SNE clusterings across different attack methods.
Essentially, \systemname\ produces consistent
t-SNE graphs across clean and adversarial examples, strongly demonstrating its ability in generalization. This is in stark contrast to all other methods that deteriorate badly 
when encountering a stronger attack. The consistent embeddings across different examples indicates that \systemname\ successfully trains a robust classifier that generates (attack) domain-invariant representations.

One might question the performance of \systemname\ because the clusters do not appear well-separated in the t-SNE plots. In fact, t-SNE does not produce conclusive evidence for clustering and classification. To genuinely measure the clustering performance of the logit embeddings, we compute the silhouette coefficient~\cite{rousseeuw1989graphical} for each sample, which is defined as $s_i = \frac{b_i-a_i}{\max\{a_i, b_i\}}$, in which $i$ is the index of the sample, $a_i$ is the intra-cluster distance of the sample and $b_i$ is the nearest cluster distance. Then the silhouette score, measuring the quality of clustering and ranging from -1 to 1 (with 1 being the best), can be computed by averaging the silhouette coefficients of all samples. 
In Table~\ref{tab:silhouette}, we report the silhouette scores of logit space embeddings for different defense methods. We can see that NT achieves the highest score on clean images, SAT is the best under the FGSM attack, and \systemname\ is the best when attacked by PGD, R+FGSM, and MIM. Higher silhouette score indicates  better clustering in the logit space, thus better classification performance. The result is consistent with the accuracy reported in Table~\ref{tab:accuracy}, hence confirming the strong cross-attack defense capability of \systemname.

\begin{table}
\vspace{2ex}
    \small  
    \centering
    \setlength\tabcolsep{4 pt}
    \caption{Silhouette scores of the logit embeddings from the Fashion-MNSIT dataset.}
    \begin{tabular}{cccccc}
    \hline
      & Clean & FGSM & PGD-20/40 & R+FGSM & MIM \\
    \hline
    NT & 0.715 & -0.052 & -0.076/-0.079 & -0.043 & -0.066\\
    SAT & 0.696 & 0.331 & 0.071/0.038 & 0.430 & 0.167\\
    EAT & 0.636 & 0.289 & 0.065/0.052 & 0.361 & 0.120\\
    ATDA & 0.686 & 0.313 & 0.109/0.082 & 0.422 & 0.181\\
    A-4-A & 0.529 & 0.307 & 0.147/0.122 & 0.445 &0.204\\
    \hline
    \end{tabular}
    \label{tab:silhouette}
    \vspace{-1.5ex}
\end{table}

\subsection{Ablation Study}
To gain a better understanding of \systemname's behavior and the effects of adversarial domain adaptation in the training process, we perform the following ablation studies.

\begin{figure*}[t]
    \centering
    \begin{subfigure}{0.45\linewidth}
    	\centering
		\includegraphics[width=\linewidth]{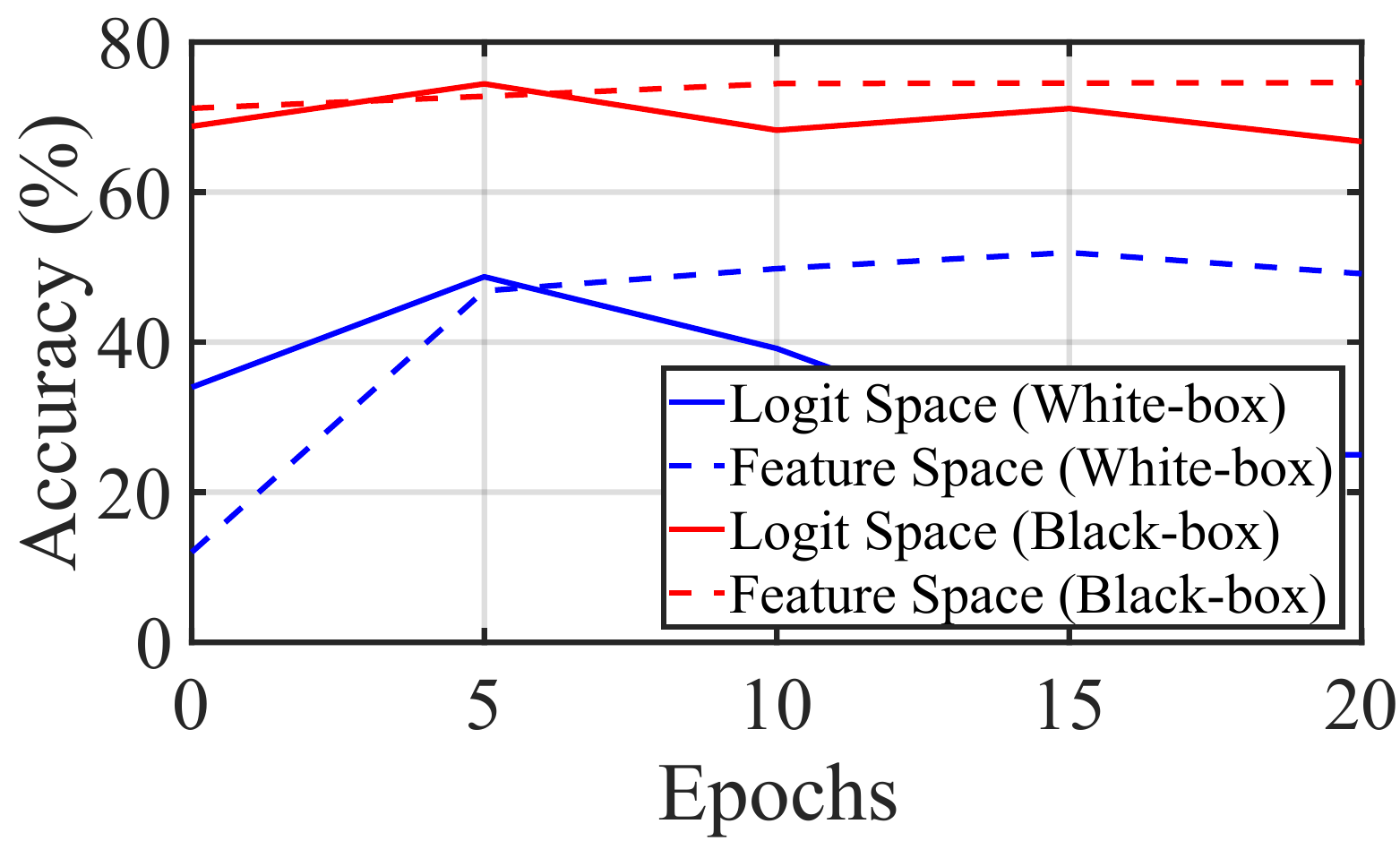}
		\caption{Fashion-MNIST.}
		\label{fig:fl_fmnist}
    \end{subfigure}
    \begin{subfigure}{0.45\linewidth}
    	\centering
		\includegraphics[width=\linewidth]{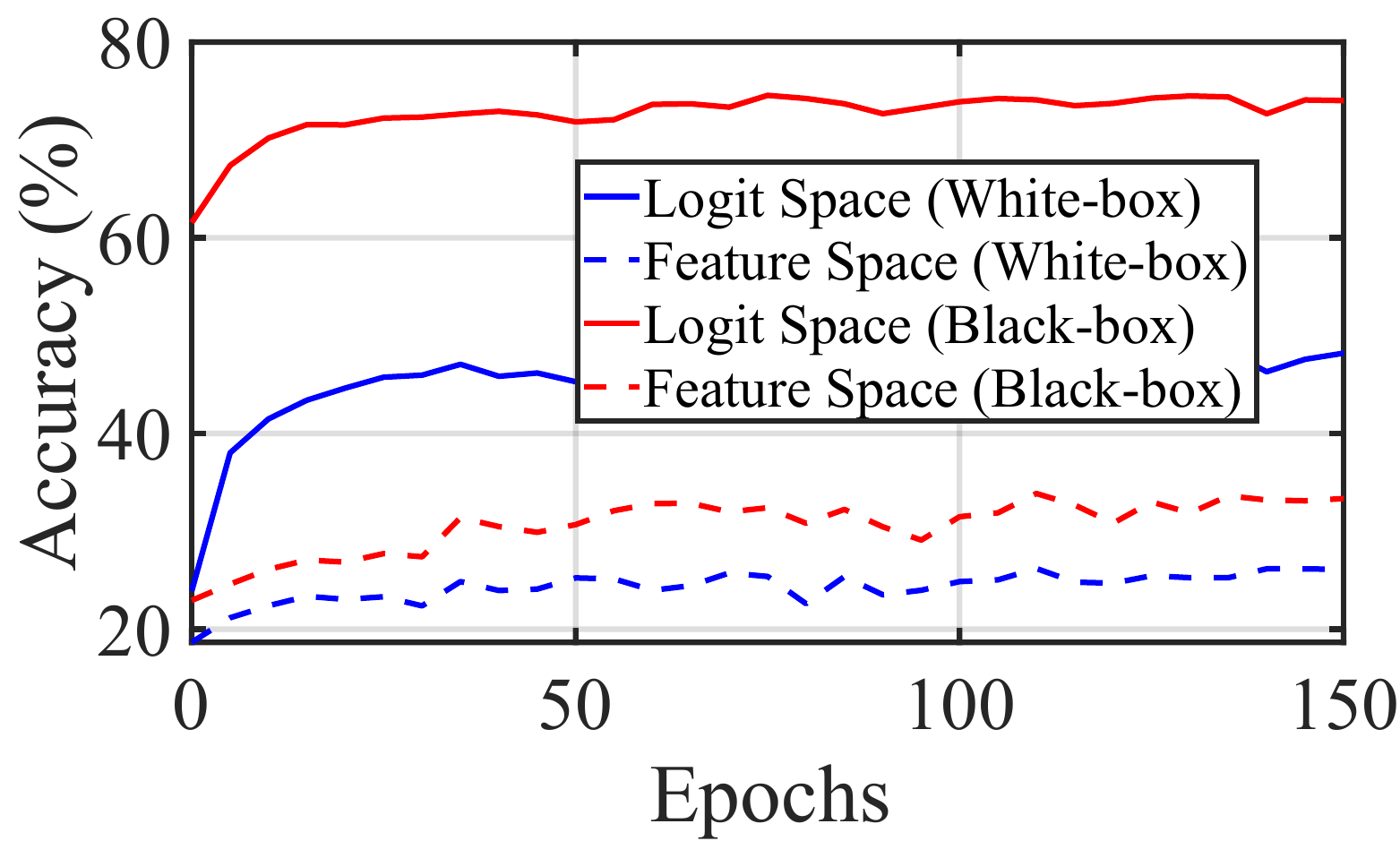}
		\caption{SVHN.}
		\label{fig:fl_svhn}
    \end{subfigure}
    \\
    \begin{subfigure}{0.45\linewidth}
    	\centering
		\includegraphics[width=\linewidth]{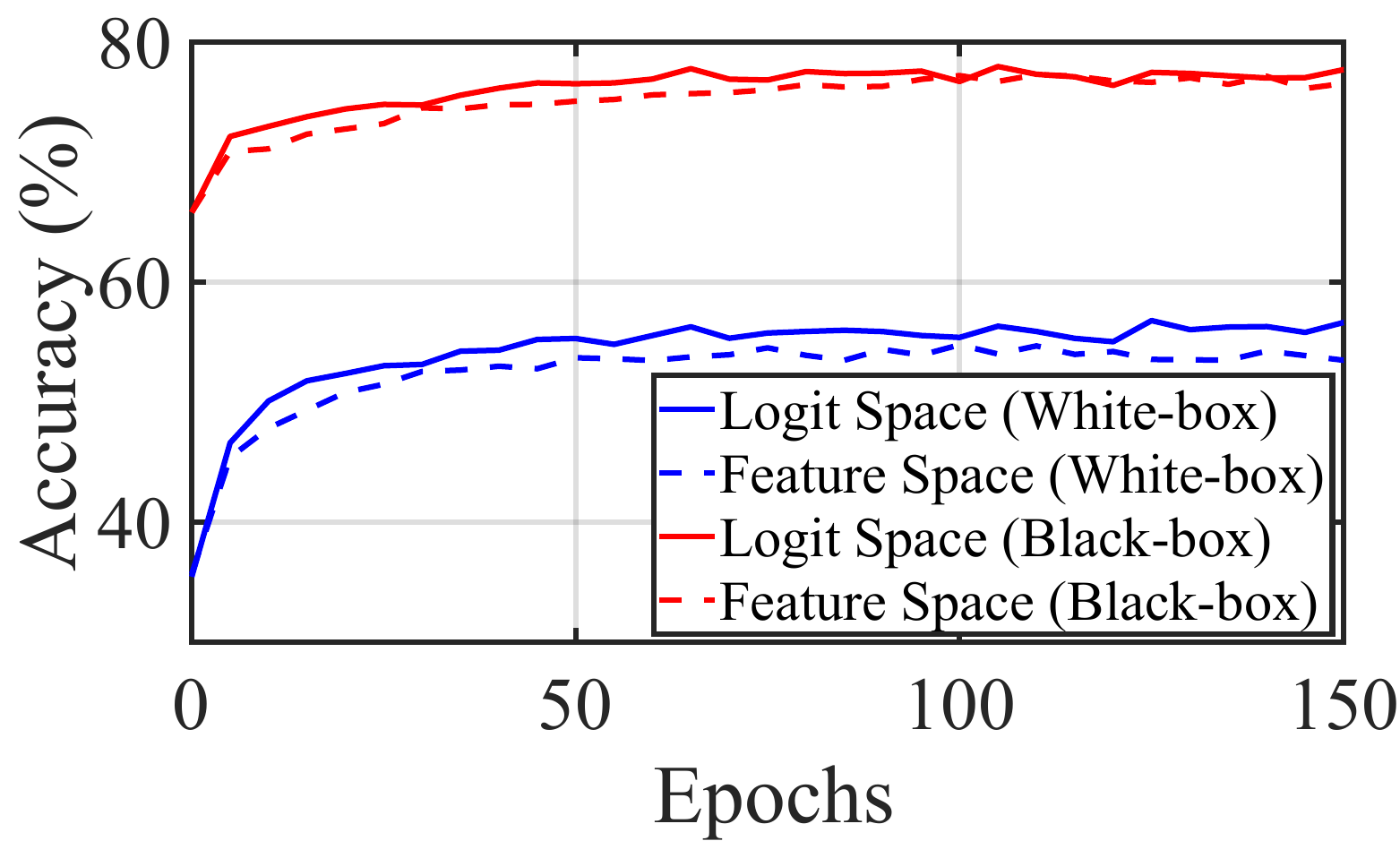}
		\caption{CIFAR-10.}
		\label{fig:fl_cifar10}
    \end{subfigure}
        \begin{subfigure}{0.45\linewidth}
    	\centering
		\includegraphics[width=\linewidth]{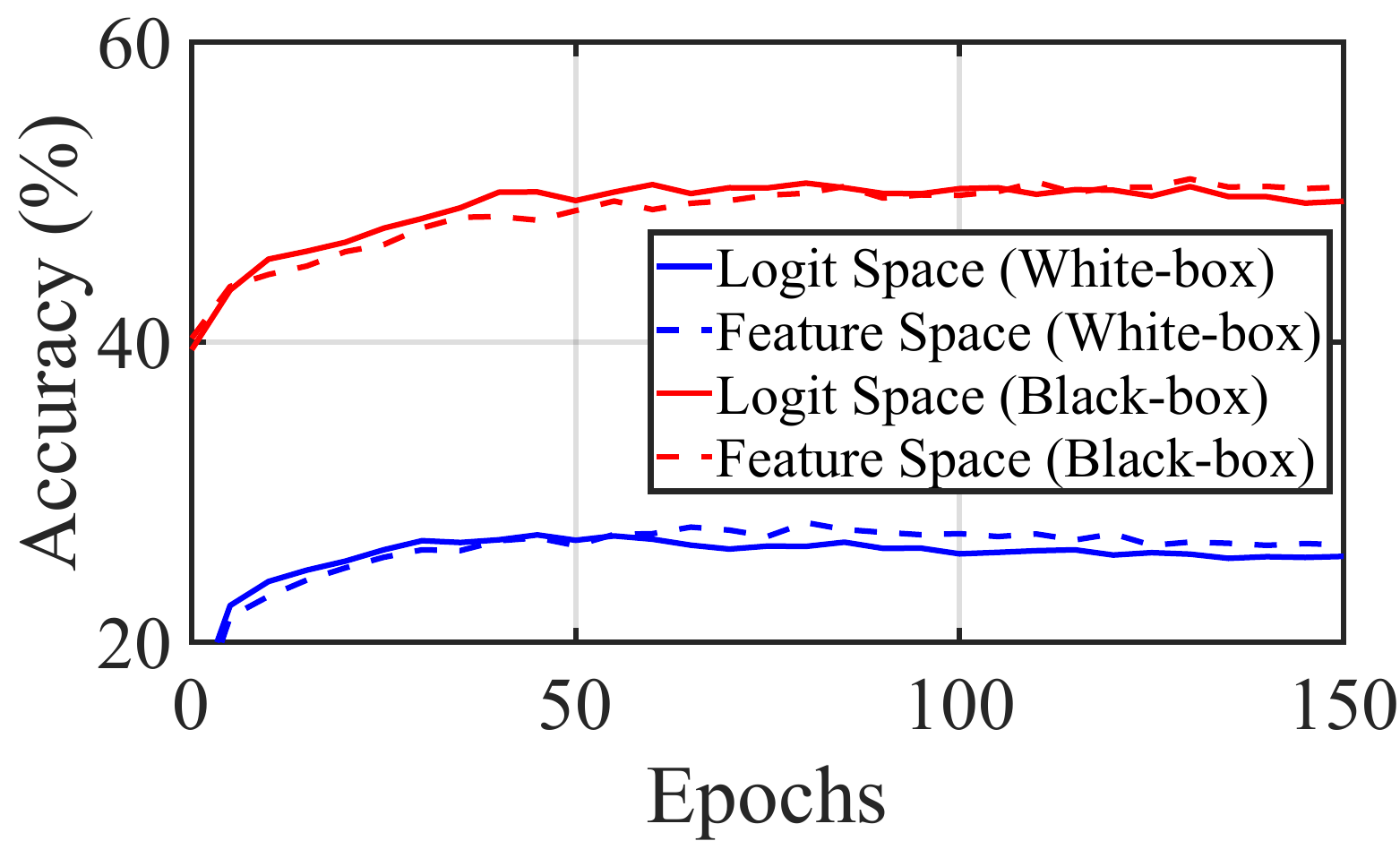}
		\caption{CIFAR-100.}
		\label{fig:fl_cifar100}
    \end{subfigure}
    \vspace{-1ex}
    \caption{Performance comparison of feature-space discriminators and logit-space discriminators.}
    \label{fig:ablation_feature}
\end{figure*}

\paragraph{Feature-Space vs.\ Logit-Space Discriminators}
As discussed in Section~\ref{ssec:representation_class}, \systemname\ uses multi-representation assisted DA, i.e., it applies discriminators on both feature space $\Phi$ and logit space $\Lambda$. We compare the performance of feature-space discriminators ($\gamma=0$ in Eqn.~\ref{eq:represetation}) and logit-space discriminators ($\beta=0$ in Eqn.~\ref{eq:represetation}) in Figure~\ref{fig:ablation_feature}; they plot how accuracy changes over training epochs of these two methods in defending PGD-20 attack. We observe that, for the Fashion-MNIST and CIFAR-100 datasets, feature-space discriminators outperform logit-space discriminators, while it is a reversed situation for the SVHN and CIFAR-10 datasets.
To combine the advantages of both representation $\Phi$ and $\Lambda$, \systemname\ adjusts the importance of the two sets of discriminators by applying different weights to the losses of the discriminators. However, since the hybrid of two sets of discriminators adds system complexity and does not significantly improve performance, we only pick whichever is the better for \systemname\ in practice.


\begin{figure*}[t]
    \centering
    \setlength\tabcolsep{4 pt}
    \begin{subfigure}{0.45\linewidth}
    	\centering
		\includegraphics[width=\linewidth]{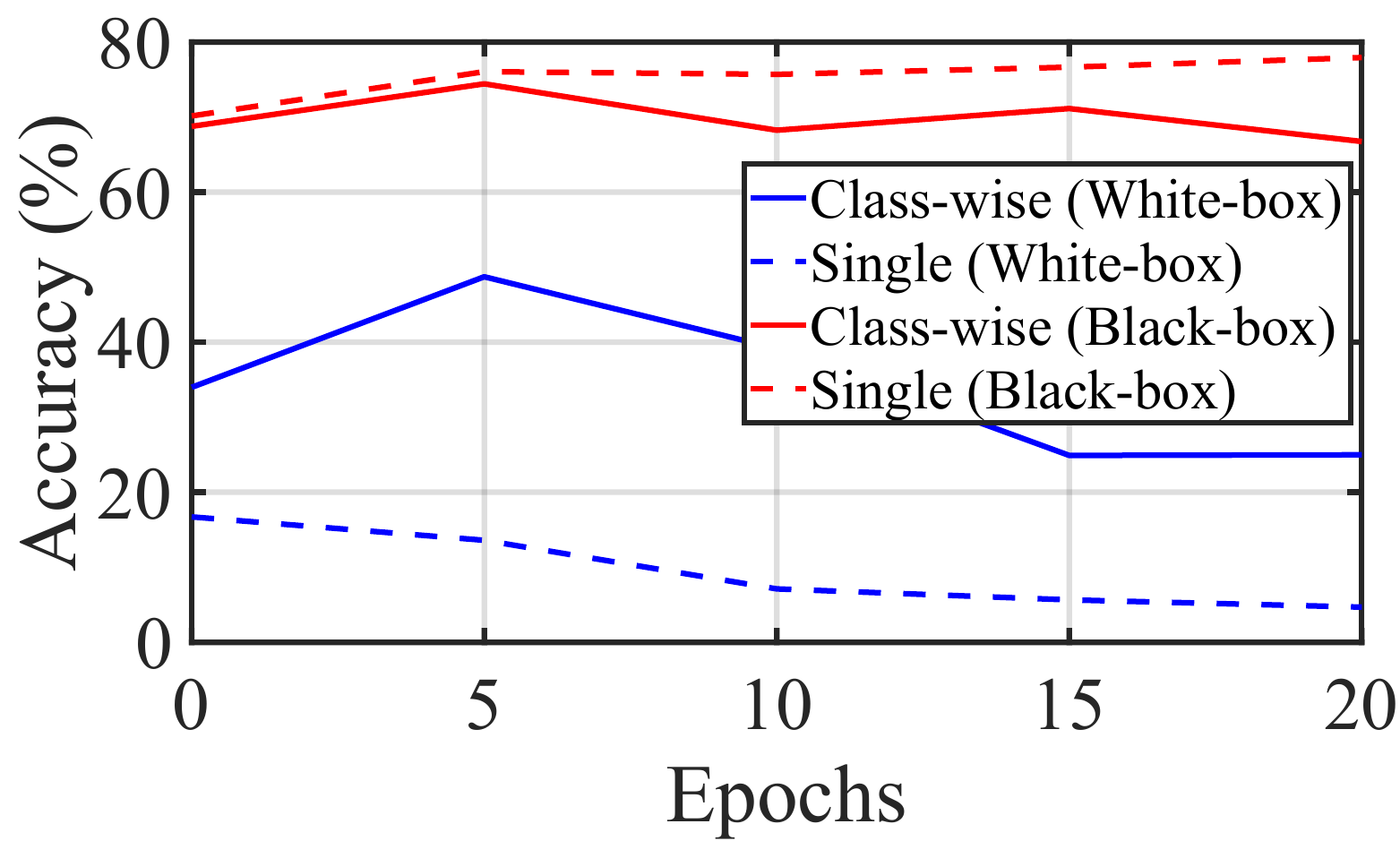}
		\caption{Fashion-MNIST.}
		\label{fig:cs_fmnist}
    \end{subfigure}
    \begin{subfigure}{0.45\linewidth}
    	\centering
		\includegraphics[width=\linewidth]{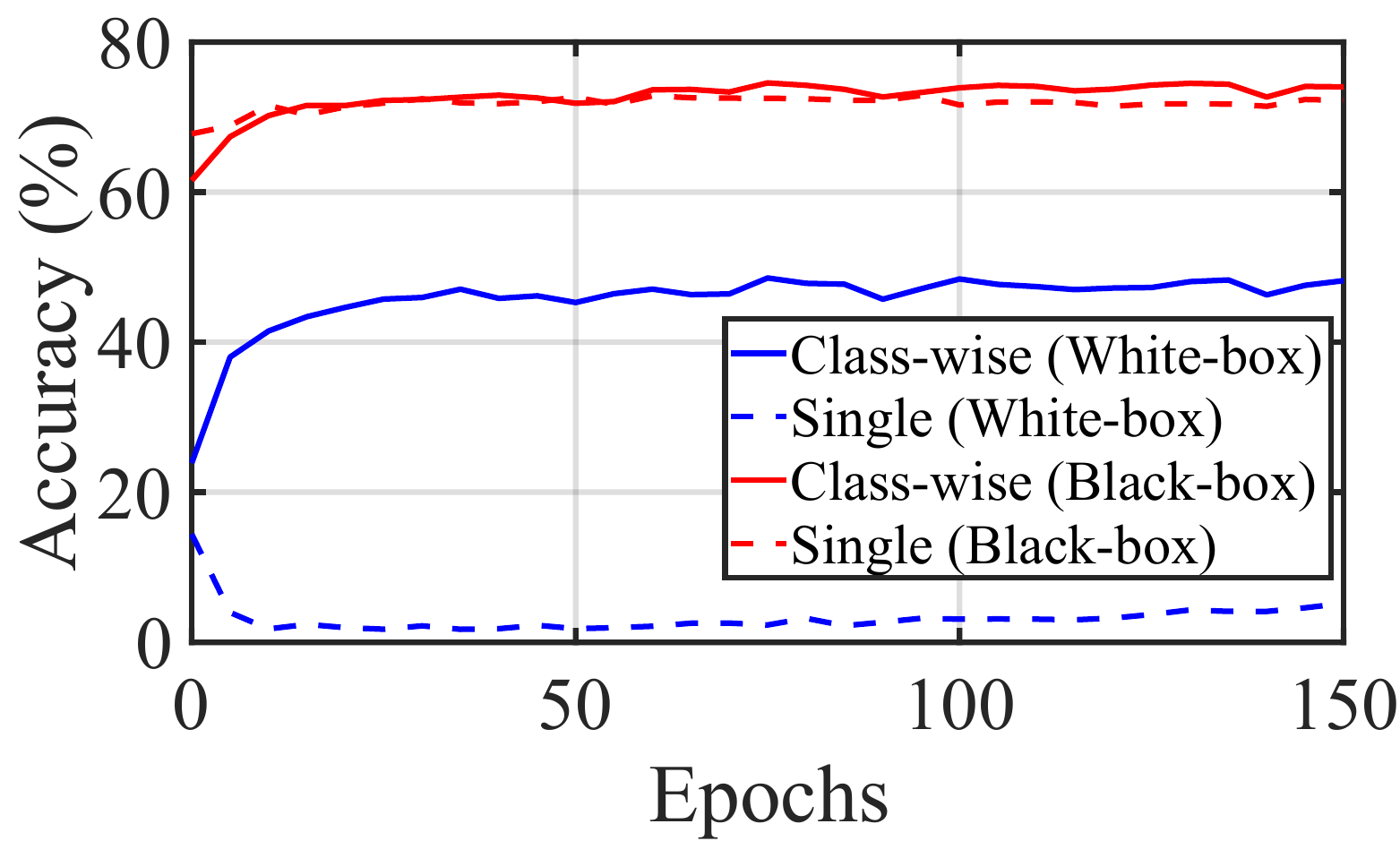}
		\caption{SVHN.}
		\label{fig:cs_svhn}
    \end{subfigure}
    \\
    \begin{subfigure}{0.45\linewidth}
    	\centering
		\includegraphics[width=\linewidth]{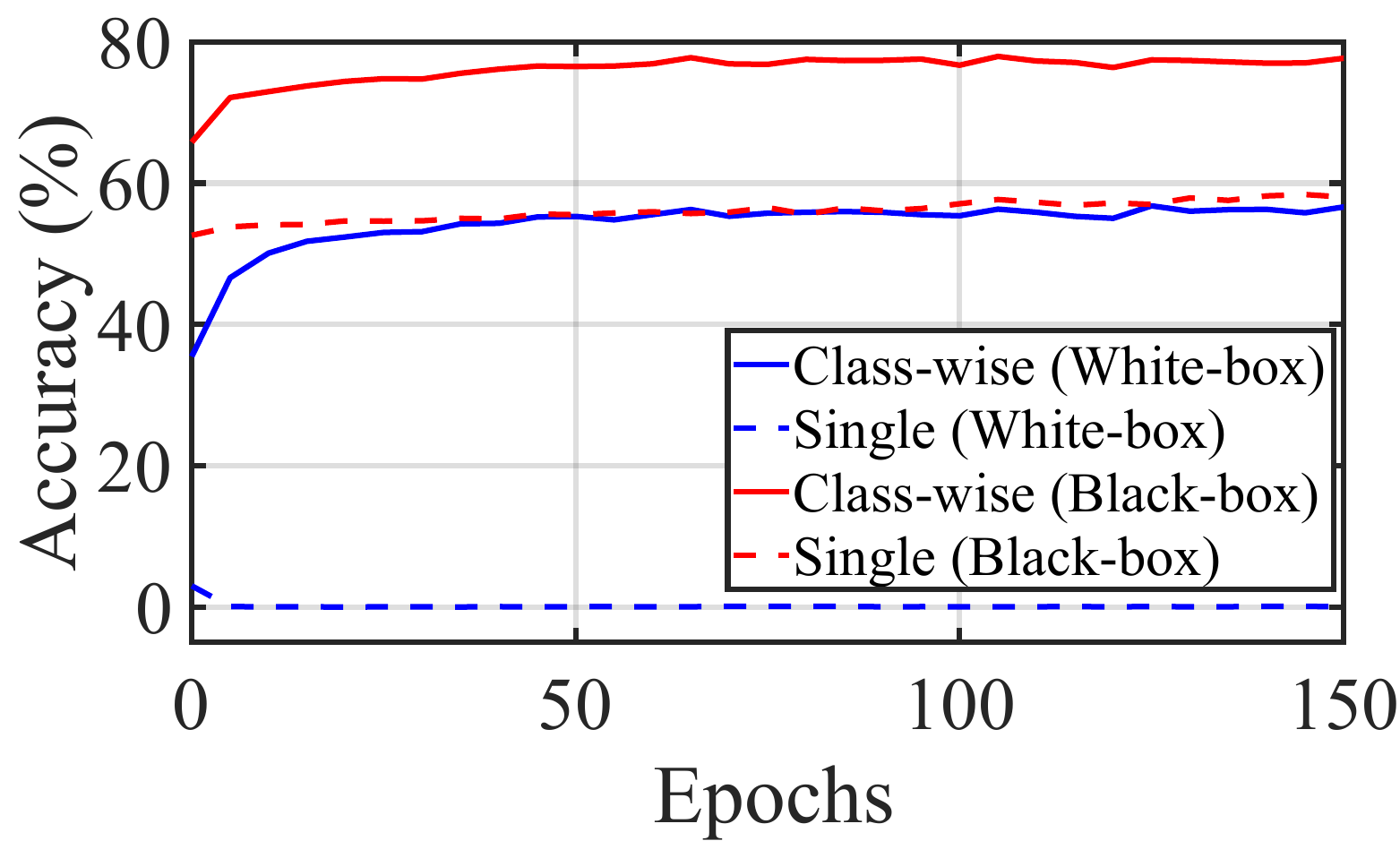}
		\caption{CIFAR-10.}
		\label{fig:cs_cifar10}
    \end{subfigure}
        \begin{subfigure}{0.45\linewidth}
    	\centering
		\includegraphics[width=\linewidth]{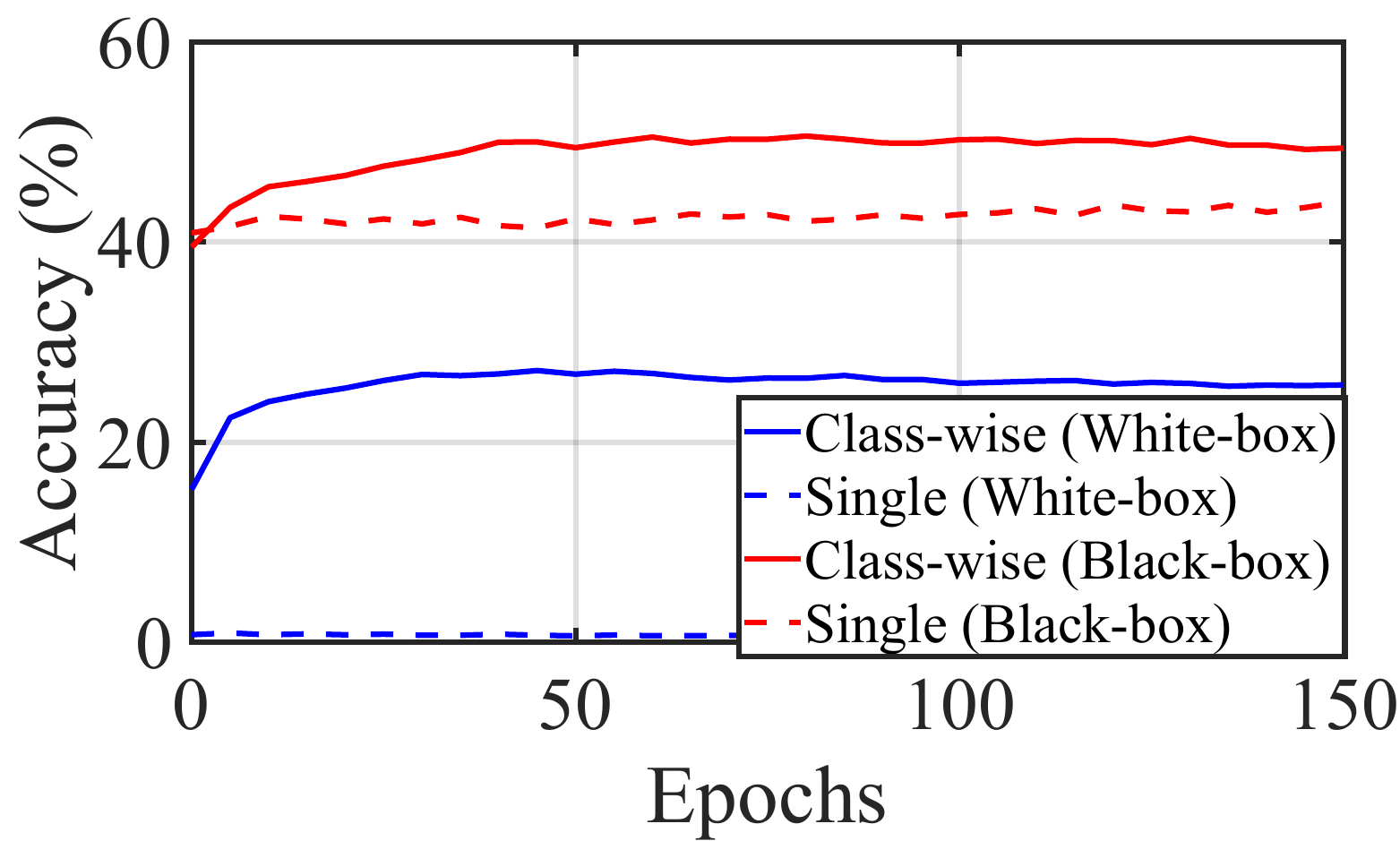}
		\caption{CIFAR-100.}
		\label{fig:cs_cifar100}
    \end{subfigure}
    \vspace{-1ex}
    \caption{Performance comparison of a single domain discriminator and class-wise discriminators.} 
    \label{fig:ablation_class_single}
    \vspace{-1.5ex}
\end{figure*}

\paragraph{Single Domain Discriminator vs.\ Class-wise Discriminators}
We conduct experiments to study the effect of class-wise discriminators. We compare the performance of class-wise discriminators with a single domain discriminator. Figure~\ref{fig:ablation_class_single} shows the accuracy of these two cases when defending PGD-20 attack. We observe that, for all white-box attacks, the accuracy of using a single domain discriminator is significantly lower than that of using class-wise discriminators, whereas a single domain discriminator has a narrow edge over class-wise discriminators for black-box attacks only on the Fashion-MNIST dataset.
The degraded performance indicates that the classifier trained with a single discriminator fails to generalize to other attack domains. This may be attributed to multiple subdomains corresponding to individual classes: a given attack might not induce a single adversarial domain, but rather separate adversarial domains corresponding to distinct classes; this renders class-wise discriminator a must.

\begin{table}[h]
    \small
    \centering
    \setlength\tabcolsep{2 pt}
    \caption{Performance of PAT, ATDA, and \systemname\, all trained by PGD. The magnitude of perturbations in $l_{\infty}$ norm is 0.1 for Fashion-MNIST (FMST), 0.02 for SVHN, and 4/255 for CIFAR-10 (C-10) and CIFAR-100 (C-100).}
    \label{tab:pgd_trained}
    \begin{tabular}{lcccccrc}
    \hline
     \multirow{2}{*}{Dataset} & \multirow{2}{*}{Defense}& Clean  & \multicolumn{4}{c}{White-Box Attack (\%)}  \\
     \cline{4-7} 
      && (\%) & FGSM & PGD-20 / 40 & R+FGSM & MIM \\
     \hline
     \multirow{3}{*}{FMST} 
     &PAT & \tb{86.86} & 77.18 & 73.40 / 73.05 & 78.12 & 74.83  \\
     &ATDA & 85.61 & 74.85 & 71.15 / 70.75 & 75.76 & 72.26 \\
     &A-4-A & 85.48 & \tb{78.88} & \tb{75.42} / \tb{75.14} & \tb{78.95} & \tb{76.46}\\
     \hline
     \multirow{3}{*}{SVHN} 
     &PAT & \tb{84.68} & \tb{58.21} & 51.62 / 50.76 & \tb{72.07} & \tb{55.57}  \\
     &ATDA & 81.67 & 54.76 & 46.31 / 45.19 & 67.56 & 51.05 \\
     &A-4-A & 80.25 & 56.84 & \tb{52.70} / \tb{51.83} & 68.35 & 55.21\\
     \hline
     \multirow{3}{*}{C-10} 
     &PAT & \tb{76.75} & 58.30 & 57.92 / \tb{58.08} & \tb{60.85} & 58.02  \\
     &ATDA & 76.72 & 58.30 & 57.86 / 57.82 & 60.25 & 57.89 \\
     &A-4-A & 75.49 & \tb{58.71} & \tb{58.16} / 57.81 & 60.03 & \tb{58.42} \\
     \hline
     \multirow{3}{*}{C-100} 
     &PAT & 50.04 & \tb{30.93} & \tb{30.39} / \tb{30.24} & \tb{32.52} & \tb{31.03}  \\
     &ATDA & \tb{53.30} & 25.18 & 24.13 / 23.48 & 27.66 & 24.59  \\
     &A-4-A & 48.00 & 30.89 & 29.82 / 29.44 & 31.99 & 30.15 \\
     \hline     
    \end{tabular}
\end{table}


\subsection{Extension to PGD-training}
\systemname\ seeks to generalize from a representative (source) adversarial domain to other related domains. 
We choose FGSM as the source domain because the attack is simple to launch, but this choice is certainly not limited to FGSM.
In this section, the source domain is crafted by the PGD-20 attack, whose perturbation in $l_{\infty}$ norm is 0.1 for Fashion-MNIST, 0.02 for SVHN, and 4/255 for CIFAR-10 and CIFAR-100. We show that \systemname\ can leverage this domain equally well in generalizing adversarial robustness.


We compare the performance of PGD-trained \systemname\ with PAT (PGD-enabled SAT) and ATDA in Table~\ref{tab:pgd_trained}. As PGD is more sophisticated and effective than FGSM, even PAT can obtain good robustness across other gradient-related attacks, but our \systemname\ still outperforms ATDA in almost all cases. These results actually make sense: the domain adaption ability \systemname\ aims for is meant to save the efforts in conducting AT for sophisticated attacks; however, if some sophisticated (e.g., PGD) has been adopted to conduct an AT, one may directly rely on the trained model to cope with other weaker attacks.


\section{Conclusion}

We have presented \systemname\ as a novel adversarial training method maintaining robustness against unseen adversarial perturbations. Leveraging the power of adversarial domain adaptation, \systemname\ can suppress the domain/attack-specific features
so as to generate a domain/attack-invariant representation. Specifically, we show that a model trained on examples crafted by single-step attacks (e.g., FGSM) can be generalized to defend against iterative attacks (e.g., PGD); 
this suggests \systemname\ is an effective alternative to the time-consuming adversarial training when dealing with advanced attacks. Compared with state-of-the-art defense methods, our extensive evaluations on four datasets show that \systemname\ achieves significantly higher cross-attack defense success rates for both white-box and black-box attacks.

\bibliographystyle{elsarticle-num} 
\bibliography{cas-refs}

\end{document}